\documentclass{article}
\usepackage{spconf,amsmath,graphicx}
\usepackage{graphicx}
\usepackage{amsmath}
\usepackage{amssymb}
\usepackage{graphicx}
\usepackage{amsmath,amssymb} % define this before the line numbering.
\usepackage[ampersand]{easylist}
\usepackage{bm}
\usepackage{pbox}
\newsavebox{\tablebox}
\usepackage{multirow}
\usepackage{booktabs}
\usepackage{threeparttable}
\usepackage{tabularx}

\title{A Baseline for Multi-Label Image Classification Using An Ensemble of Deep Convolutional Neural Networks}
\name{Qian Wang, Ning Jia, Toby P. Breckon}
\address{Department of Computer Science, Durham Univeristy, United Kingdom}
\begin{document}
%\ninept
%
\maketitle
\begin{abstract}
   Recent studies on multi-label image classification have focused on designing more complex architectures of deep neural networks such as the use of attention mechanisms and region proposal networks. Although performance gains have been reported, the backbone deep models of the proposed approaches and the evaluation metrics employed in different works vary, making it difficult to compare fairly. Moreover, due to the lack of properly investigated baselines, the advantage introduced by the proposed techniques are often ambiguous. To address these issues, we make a thorough investigation of the mainstream deep convolutional neural network architectures for multi-label image classification and present a strong baseline. With the use of proper data augmentation techniques and model ensembles, the basic deep architectures can achieve better performance than many existing more complex ones on three benchmark datasets, providing great insight for the future studies on multi-label image classification.
\end{abstract}
\section{Introduction}

% the introduction of multi-label image classification
Multi-label image classification has been a hot topic in computer vision community. Its extensive applications include but are not limited to image retrieval, automatic image annotation, web image search and image tagging \cite{guillaumin2009tagprop,chen2013fast,zhou2011hybrid,zhang2016fast,wang2017multi2}.

% the use of deep CNN in multi-label image classification
The abundant labelled data (e.g. ImageNet \cite{russakovsky2015imagenet}) and advanced computational hardware have promoted the development of deep convolutional neural network (CNN) based methods on single-label image classification \cite{krizhevsky2012imagenet, he2016deep}.  Recently, such successful models have been extended to multi-label classification tasks with promising performance reported by \cite{wang2016cnn,wang2017multi,li2017improving,zhu2017learning,yeh2017learning,zhang2018kill,zhang2018multi}, proving that CNN models are capable of handling this challenging and more general problem. However, due to the varying backbones \cite{zhang2018multi,chen2018multi} employed in the deep models, the achieved performance cannot be directly compared with each other. In addition, the lack of thoroughly investigated baselines of these deep CNN models hinders an explicit evaluation of the benefit brought by advanced frameworks specially designed for multi-label image classification.

% Our work in this paper and contributions
To address the aforementioned issues, we present a thorough investigation on different baseline deep CNN models for multi-label image classification. We focus on two state-of-the-art deep CNN architectures (i.e., VGG16 \cite{simonyan2015very} and ResNet101\cite{he2016deep}) as they have been widely employed in multi-label image classification \cite{zhu2017learning,zhang2018multi}. We evaluate the models by taking advantage of varying data augmentation techniques and model ensemble, surprisingly achieving comparable or superior performance on three benchmark datasets than the state-of-the-art results achieved by more complex models. 
%Do not use any additional Latex macros.

The contributions of this work are summarized as follows:
\begin{itemize}
	\item{We investigate the impacts of varying image sizes and data augmentation techniques including ``mixup" which has not been employed in multi-label image classification.}
	\item{We use score level fusion to investigate the complementarity of different models and point out possible directions for future model design.}
	\item{We present a strong baseline for multi-label image classification with performance comparable with state-of-the-art on three benchmark datasets.}
\end{itemize}
%------------------------------------------------------------------------- 
\section{Related Work}
\label{sect:related}

Impressive progress on multi-label image classification has been made by using deep convolutional neural networks. Wang et al. \cite{wang2016cnn} propose a CNN-RNN framework to explore label co-occurrence using the long-short term memory (LSTM). Although VGG16 was employed as a visual feature extractor, the model capacity was not fully exploited by fine-tuning the parameters. 
Zhang et al. \cite{zhang2018multi} extend the idea by improving the component CNN. They propose a regional latent semantic dependencies (RLSD) model for multi-label image classification, which focuses on small objects in  multi-label images by generating subregions that potentially contain multiple objects and visual concepts. An LSTM based model is employed to generate multiple labels. 
Recently, attention mechanisms have been introduced to deep neural networks for multi-label image classification. It aims to explicitly or implicitly extract multiple visual representations from a single image characterizing different associated labels \cite{wang2017multi,zhu2017learning}. 
The advantage of combining multi-scale input images for multi-label image classification has been proved in \cite{wang2016beyond, durand2017wildcat} by employing varying fusion approaches. 

Although improved performance has been reported by introducing more advanced frameworks, we notice that the performance of those proposed methods has marginal gains towards the standard (``vanilla") deep models, and the training techniques employed in different works vary. Therefore it is necessary to set up a uniform baseline for comparison. 
%===========================================================
\section{Method}
\label{sect:method}
We present the methods used to produce the strong baseline performance in this section. We first formulate the multi-label image classification problem. Subsequently, we describe the adapted deep convolutional neural networks for multi-label classification, as well as the essential data augmentation techniques for training an improved deep model. Finally, A simple yet effective model ensemble approach is introduced to investigate the complementarity of different models.

\begin{figure}
	\label{fig:framework}
	\includegraphics[width=0.5\textwidth]{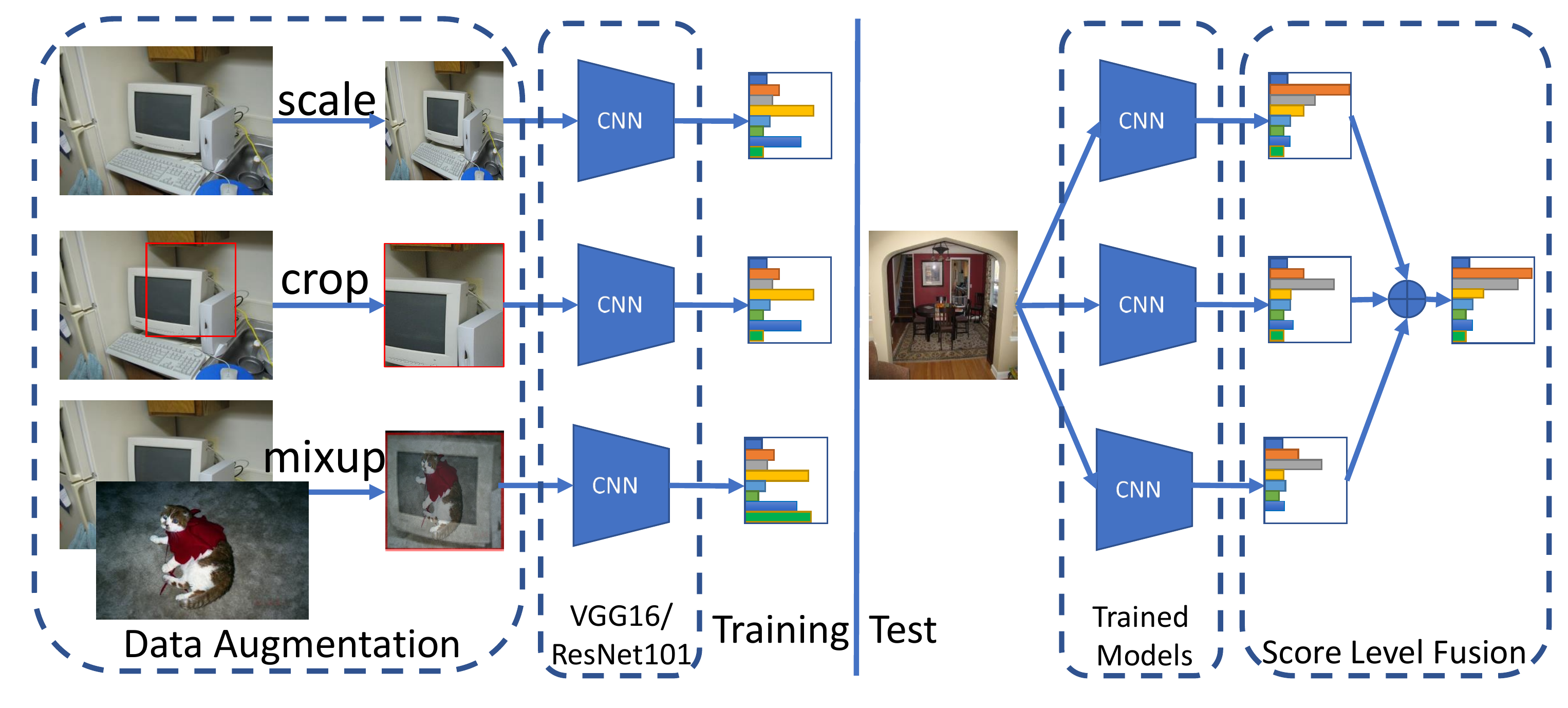}
	\caption{An illustration of the employed framework. Varying image scales and data augmentation techniques are used during training which results in diverse trained models. These models are used for testing individually and combined by score level fusion.}
\end{figure}

%------------------------------------------------------------------------- 
\subsection{Problem Formulation}
Assume we have a set of training examples $\mathcal{D}=\{(\bm{x}_i,\bm{y}_i)\},i=1,2,...,n$, where $\bm{x}$ is an image, $\bm{y}\in \{0,1\}^C$ is the corresponding label vector, $n$ and $C$ are the numbers of training images and associated class labels respectively, the element values of zeros and ones in the label vector $\bm{y}$ denote the absence and presence of the corresponding concepts in the image. The objective of multi-label image classification is to learn a model from the training data $\mathcal{D}$, such that for a given test image $\hat{\bm{x}}$ we can use the learned model to predict its label vector $\hat{\bm{y}}$. In practice, most parametric models do not directly output a binary vector $\hat{\bm{y}}$, instead they predict a score vector $\hat{\bm{s}} = f(\hat{\bm{x}};\Theta) \in \mathbb{R}^C$ indicating the confidence of presence for each label. $\hat{\bm{y}}$ can be derived from $\hat{\bm{s}}$ by setting a threshold confidence or the number of positive labels \cite{li2017improving}.
%------------------------------------------------------------------------- 
\subsection{Base Model} 
% deep models
Deep convolutional neural networks can be used to implement the model $f(\bm{x};\Theta)$ for multi-label image classification with an image $\bm{x}$ as the input and a $C$-dimensional score vector $\bm{s}$ as the output. In contrast to the traditional multi-label classification approaches, deep models integrate the feature extraction and classification in a single framework, enabling end-to-end learning. More importantly, state-of-the-art deep CNN models are able to learn high-level visual representations and approximate very complex learning systems.

\textbf{Model adaptation:}
we focus on two deep CNN architectures which have been used in multi-label image classification: VGG16 \cite{simonyan2015very} and Resnet-101 \cite{he2016deep} to which two changes have been made in this study. First, we apply an adaptive pooling layer to the last convolutional feature maps such that different input sizes can be handled within the same architecture. Second, the final output layer for single-label classification in the original model is simply replaced with a fully connected layer in which the number of neurons is set as $C$ (i.e. the number of concerned class labels).

% loss functions
\textbf{Loss function:}
We use the cross-entropy loss for model training. For a training example $(\bm{x}_i,\bm{y}_i)$ and its predicted score vector $\bm{s}_i = f(\bm{x}_i;\Theta)$, the loss can be computed by the following equation:
\begin{equation}
\label{eq:loss}
L(\bm{s}_i, \bm{y}_i) = -\sum_{j=1}^C \big(y_{ij}\cdot\log (\sigma(s_{ij})) + (1-y_{ij}) \cdot \log (1-\sigma(s_{ij}))\big)
\end{equation}
where $y_{ij}$ is the $j$-th element of the ground truth label vector $\bm{y}_i$, $s_{ij}$ is the $j$-th element of the predicted score vector $\bm{s}_i$, and $\sigma(\cdot)$ is the sigmoid function $\sigma(x) = 1/(1+\exp(-x))$. 

\textbf{Data Augmentation:} 
we aim to investigate how different data augmentation techniques affect the multi-label image classification. This is non-trivial since some commonly adopted data augmentation techniques such as random cropping will change the semantics in the original image. For example, a random cropping of a multi-label image might result in image patches not containing all the objects in the original image thus it is questionable whether they are still applicable to multi-label classification. 

Apart from the conventional data augmentation techniques, we also adapt the \textit{mixup} \cite{zhang2018mixup,inoue2018data} method to further increase data variability. Specifically, we randomly select two samples $(\bm{x}_i,\bm{y}_i)$ and $(\bm{x}_j,\bm{y}_j)$ from the mini-batch (the samples in a mini-batch could be image patches cropped from the original images, resized to the same size). The mixed sample $(\bm{x},\bm{y})$ can be created in the following way:
\begin{align}
\label{eq:mixup}
\bm{x} = (\bm{x}_i + \bm{x}_j)/2,
\notag \\
\bm{y} = \bm{y}_i \lor \bm{y}_j, \mspace{36mu}
\end{align}
where the mixed image $\bm{x}$ is created by a pixel-wise average on two original images and the corresponding label vector $\bm{y}$ is obtained by an element-wise logical \textbf{OR} operation on $\bm{y}_i$ and $\bm{y}_j$. 
During training, the \textit{mixup} is alternately enabled and disabled for every epoch as suggested in \cite{inoue2018data}. We investigate the \textit{mixup} technique due to the fact that it expands the target label space significantly which are quite different from other traditional data augmentation techniques.
%------------------------------------------------------------------------- 
\subsection{Model Ensemble}
\label{sect:ensemble}
We explore the complementarity of models learned in different settings by a simple score level fusion which is employed during the testing phase. Suppose we have $m$ score matrices $\bm{S}_i, i=1,2,...,m$ predicted by $m$ base models, the fused score matrix $\bm{S}^{fusion}$ can be computed as follows:
\begin{equation}
\label{eq:fusion}
\bm{S}^{fusion} = \frac{1}{m} \sum_{i=1}^m \bm{S}_i.
\end{equation}

We investigate two approaches to promote the diversity of base models for better ensemble performance. Firstly, we combine models trained with different input image sizes and denote this ensemble as a \textit{multi-scale ensemble} or \textit{ScaleEn}. The complementarity of multi-scale input images has been explored before \cite{wang2016beyond, durand2017wildcat} but in  different ways. Secondly, we combine models trained with different data augmentation techniques. Using varying data augmentation results in different training data distribution thus diversifies the learned models. We denote this ensemble as a \textit{distribution ensemble} or \textit{DistrEn}.
%------------------------------------------------------------------------- 
%===========================================================
\section{Experiments and Results}
\label{sect:datasets}
In this section, we describe our experiments on three benchmarks and report the experimental results. We introduce the datasets used in our experiments and the implementation details of the deep model training in the first two subsections respectively, then experimental results are presented in the last subsection.
%------------------------------------------------------------------------- 
\subsection{Dataset}

\begin{table}[!ht]
	\centering
	{%\centering
		\caption[]{A summary of datasets used in our experiments.\\
		}
		\label{table:datasets}
		\begin{lrbox}{\tablebox}
			\begin{tabular}{c|c|c|c}
				\hline
				\textbf{Dataset} & \textbf{\# Labels} & \textbf{\# Training Images} & \textbf{\# Test Images}\\
				\hline
				NUS-WIDE \cite{chua2009nus} & 81 & 100,893 & 67,742 \\ 
				MS-COCO \cite{lin2014microsoft} & 80 &82,081 & 40,137\\
				VOC2007 \cite{everingham2010pascal} & 20 &5,011& 4,952 \\
				\hline
			\end{tabular}
		\end{lrbox}
		\scalebox{0.70}{\usebox{\tablebox}}
	}
\end{table}

We use three benchmark datasets for multi-label image classification in our experiments, i.e., NUS-WIDE \footnote{Many image urls are not valid now, as a result, our experiments are actually conducted on a subset of the original dataset. } \cite{chua2009nus}, MS-COCO \cite{lin2014microsoft} and VOC 2007 \cite{everingham2010pascal}. A summary of three datasets is presented in Table \ref{table:datasets}. 
%------------------------------------------------------------------------- 
\subsection{Implementation}
\label{sect:implementation}
All the deep CNN models used in our experiments are implemented in PyTorch \footnote{https://github.com/hellowangqian/multi-label-image-classification} \cite{paszke2017automatic}. We use the model weights pre-learned on the ImageNet \cite{russakovsky2015imagenet} for single-label image classification as the initialization and fine-tune the weights of all layers. We use the stochastic gradient descent (SGD) optimizer for model training with an initial learning rate of 0.1 for the fully connected layer(s) and 0.01 for convolutional layers. The learning rate decays to one tenth after 20 epochs. We stop training after 40 epochs. The batch size is set 16 in all experiments.
%------------------------------------------------------------------------- 
\subsection{Experiments and Results}
Table \ref{table:main} shows experimental results on three datasets. For each dataset, we use two deep models (i.e., VGG16 and ResNet101 denoted as V and R respectively in the table). We first investigate the impact of input image size. Three sizes (i.e., 384, 448 and 512) are employed for each experimental setting. The results in Table \ref{table:main} indicate different input image sizes  do not affect the results except on the MS-COCO dataset where a larger input image size generally performs better. One possible explanation is that images in the MS-COCO dataset have larger sizes than those in the other two datasets, as a results, rescaling them to a small size (e.g., $384\times 384$) causes information loss. 

 We investigate the effectiveness of data augmentation by using three models. The first model (M1) uses only random flipping for data augmentation which has also been used in all experiments. The second model (M2) uses  \textit{randomly resized cropping} for data augmentation which randomly rescale and crop the image \footnote{See the implementation of  \textit{transforms.RandomResizedCrop} in PyTorch.}. The \textit{mixup} strategy \cite{zhang2018mixup} is employed in the third model (M3). Experimental results in Table \ref{table:main} show superior performance when using data augmentation strategies (e.g., M2 and M3 perform better than M1 except on the VOC2007 dataset when ResNet101 is used where the model without any data augmentation performs the best). By comparing the performance of M2 and M3, we find that \textit{mixup} does not improve the performance in most cases. However, the models learned with \textit{mixup} provides complementary information to those learned without it. This can be verified by our model ensemble results shown in Table \ref{table:resCmp}.

\begin{table}[!h]
	\centering
	{%\centering
		\caption[]{Experimental results on three benchmark datasets.\\
			\footnotesize The precision, recall and F$_1$ are based on top-3 predictions without any threshold conditions.(Notations: DS--DataSet, BM--Base Model, V--VGG16, R--ResNet101, Size--input image Size, mAP--mean Average Precision, L-P/R/$F_1$ -- Label centric Precision/Recall/F1 score, O-P/R/$F_1$ -- Overall Precision/Recall/$F_1$ score, M1 -- Model with random flipping, M2 -- Model with random cropping, M3 -- Model with mixup.)
			
		}
		\label{table:main}
		\begin{lrbox}{\tablebox}
			\begin{tabular}{c|c|c|c|c|ccc|ccc}
				\hline
				\textbf{DS}&\textbf{BM}&\textbf{M}&\textbf{Size}& \textbf{mAP}  & \textbf{L-P} & \textbf{L-R} & \textbf{L-F$_1$} & \textbf{O-P} & \textbf{O-R} & \textbf{O-F$_1$}\\ 
				
				\hline
				\multirow{18}{*}{NUS} 
				&\multirow{9}{*}{V} &\multirow{3}{*}{M1} 
				  & 384  & 55.8 &37.7 & 57.3 & 42.2 & 54.0 & 66.5 & 59.6\\
				&&& 448  & 55.5 &37.1 & 57.0 & 41.6 & 53.9 & 66.3 & 59.5\\
				&&& 512  & 56.5 &39.4 & 57.0 & 43.1 & 54.3 & 66.9 & 59.9\\
   %&&& ScaleEn-M1(123)  & 58.5 &39.2 & 58.2 & 43.1 & 55.0 & 67.7 & 60.7\\
   %&&& ScaleEn-M1( 23)  & 57.7 &38.9 & 57.8 & 42.8 & 54.7 & 67.3 & 60.4\\
   
				\cline{3-11}	
				&& \multirow{3}{*}{M2}
				  & 384  & 58.9 &46.6 & 55.0 & 46.1 & 55.9 & 68.9 & 61.7\\ 
				&&& 448  & 59.0 &45.7 & 55.1 & 46.6 & 55.9 & 68.9 & 61.7\\	
				&&& 512  & 58.8 &45.9 & 55.0 & 46.5 & 55.9 & 68.8 & 61.7\\
	%&&& ScaleEn-M2(123) & 59.2 &46.6 & 55.3 & 47.0 & 56.1 & 69.1 & 61.9\\
	%&&& ScaleEn-M2( 23) & 59.0 &45.7 & 55.1 & 46.7 & 56.0 & 69.0 & 61.8\\
	
				\cline{3-11}	
				&& \multirow{3}{*}{M3}
				  & 384  & 58.8 &46.5 & 54.9 & 47.0 & 55.9 & 68.8 & 61.7\\ 
				&&& 448  & 58.5 &46.5 & 54.4 & 46.7 & 55.9 & 68.8 & 61.7\\	
				&&& 512  & 58.3 &46.4 & 54.3 & 46.6 & 55.9 & 68.8 & 61.6\\
	%&&& ScaleEn-M3(123) & 59.1 &47.2 & 54.9 & 47.2 & 56.1 & 69.0 & 61.9\\
	%&&& ScaleEn-M3( 23) & 58.6 &46.8 & 54.6 & 46.9 & 56.0 & 68.9 & 61.8\\
	
		%DistrEn123-384 & 59.7 &42.0 & 57.3 & 44.6 & 55.8 & 68.7 & 61.6\\
		%DistrEn 23-384 & 59.6 &47.2 & 55.3 & 47.3 & 56.2 & 69.1 & 62.0\\
		%DistrEn123-448 & 59.6 &42.6 & 57.2 & 44.8 & 55.8 & 68.7 & 61.6\\
		%DistrEn 23-448 & 59.5 &47.1 & 55.3 & 47.3 & 56.2 & 69.2 & 62.0\\
		%DistrEn123-512 & 59.9 &44.1 & 56.8 & 45.7 & 56.0 & 68.9 & 61.8\\
		%DistrEn 23-512 & 59.3 &47.0 & 55.0 & 47.0 & 56.2 & 69.1 & 62.0\\
				\cline{2-11}
				
				&\multirow{9}{*}{R} &\multirow{3}{*}{M1} 
				  & 384 & 59.0 &44.6 & 56.8 & 45.0 & 56.3 & 69.3 & 62.1\\
				&&& 448 & 59.2 &44.1 & 57.0 & 45.9 & 56.4 & 69.4 & 62.2\\
				&&& 512 & 59.2 &44.1 & 57.0 & 45.9 & 56.4 & 69.4 & 62.2\\
   %&&& ScaleEn-M1(123) & 59.8 &45.2 & 57.6 & 46.1 & 56.7 & 69.8 & 62.5\\
   %&&& ScaleEn-M1( 23) & 59.6 &44.7 & 57.7 & 46.8 & 56.6 & 69.6 & 62.4\\   
				\cline{3-11}	
				&& \multirow{3}{*}{M2}
				  & 384 & 60.8 &46.1 & 60.6 & 48.4 & 56.1 & 69.0 & 61.9\\
				&&& 448 & 60.8 &45.8 & 60.6 & 49.2 & 56.2 & 69.2 & 62.0\\
				&&& 512 & 60.6 &45.4 & 60.9 & 49.0 & 56.1 & 69.0 & 61.9\\
   %&&& ScaleEn-M2(123) & 61.4 &46.0 & 61.0 & 49.5 & 56.3 & 69.4 & 62.2\\
   %&&& ScaleEn-M2( 23) & 61.2 &46.1 & 61.0 & 49.4 & 56.3 & 69.3 & 62.1\\
				\cline{3-11}	
				&& \multirow{3}{*}{M3}
				  & 384 & 60.3 &45.2 & 60.1 & 48.8 & 56.2 & 69.2 & 62.0\\ 
				&&& 448 & 60.5 &45.1 & 60.2 & 48.8 & 56.2 & 69.2 & 62.0\\
				&&& 512 & 60.1 &46.1 & 59.5 & 49.0 & 56.2 & 69.2 & 62.0\\
   %&&& ScaleEn-M3(123) & 61.7 &46.9 & 60.5 & 49.9 & 56.7 & 69.7 & 62.5\\
   %&&& ScaleEn-M3(123) & 61.4 &46.9 & 60.4 & 49.7 & 56.6 & 69.7 & 62.5\\
		%DistrEn(123)-384 & 62.2 &46.7 & 60.4 & 49.3 & 56.8 & 69.9 & 62.7\\
		%DistrEn(23) -384 & 62.0 &47.5 & 61.1 & 49.4 & 56.6 & 69.7 & 62.5\\
		%DistrEn(123)-448 & 62.1 &46.7 & 60.5 & 49.4 & 56.8 & 70.0 & 62.7\\
		%DistrEn(23) -448 & 62.0 &46.8 & 61.1 & 49.9 & 56.7 & 69.8 & 62.6\\
		%DistrEn(123)-512 & 61.9 &46.9 & 60.8 & 49.7 & 56.8 & 69.9 & 62.7\\
		%DistrEn(23) -512 & 61.3 &46.8 & 60.8 & 49.8 & 56.5 & 69.5 & 62.3\\
				\hline
				\hline		
						
				\multirow{18}{*}{COCO} 
				&\multirow{9}{*}{V} &\multirow{3}{*}{M1}  
				  & 384  & 71.6  & 55.2 & 61.6 & 56.5 & 62.6 & 64.7 & 63.6\\
				&&& 448  & 71.4  & 54.7 & 61.6 & 56.2 & 62.5 & 64.6 & 63.5\\
				&&& 512  & 72.3  & 55.3 & 62.0 & 56.8 & 63.0 & 65.1 & 64.0\\
    %&&& ScaleEn-M1(123) & 74.3  & 56.9 & 63.3 & 58.0 & 64.1 & 66.2 & 65.2\\
    %&&& ScaleEn-M1( 23) & 73.5  & 56.2 & 62.9 & 57.5 & 63.7 & 65.8 & 64.8\\    
				\cline{3-11}	
				&& \multirow{3}{*}{M2}
				  & 384 & 75.2  & 63.2 & 62.5 & 62.7 & 64.6 & 66.7 & 65.7\\ 
				&&& 448 & 75.8  & 63.1 & 63.3 & 61.7 & 65.0 & 67.1 & 66.0\\	
				&&& 512 & 76.0  & 63.3 & 63.4 & 62.5 & 65.0 & 67.2 & 66.1\\
    %&&& ScaleEn-M2(123)& 76.9  & 64.9 & 63.8 & 62.5 & 65.5 & 67.7 & 66.6\\    
    %&&& ScaleEn-M2( 23)& 76.7  & 64.4 & 63.7 & 62.3 & 65.4 & 67.6 & 66.5\\    
    			\cline{3-11}	
				&& \multirow{3}{*}{M3}
				  & 384 & 75.1  & 64.2 & 62.3 & 63.1 & 64.7 & 66.9 & 65.8\\ 
				&&& 448 & 75.9  & 64.3 & 62.6 & 63.4 & 65.0 & 67.1 & 66.1\\	
				&&& 512 & 75.8  & 64.3 & 62.6 & 63.4 & 65.0 & 67.1 & 66.1\\
    %&&& ScaleEn-M3(123)& 76.5  & 65.2 & 63.0 & 64.0 & 65.4 & 67.5 & 66.4\\     
	%&&& ScaleEn-M3( 23)& 76.4  & 64.9 & 62.9 & 63.9 & 65.3 & 67.4 & 66.4\\	
   	%DistrEn-384(123)   & 75.4  & 59.6 & 63.3 & 59.9 & 64.7 & 66.8 & 65.7\\   	
	%DistrEn-384(23)    & 76.1  & 64.7 & 63.0 & 63.7 & 65.2 & 67.3 & 66.2\\	
   	%DistrEn-448(123)   & 75.9  & 59.9 & 63.7 & 60.3 & 64.9 & 67.0 & 65.9\\   	
	%DistrEn-448(23)    & 76.8  & 65.5 & 63.5 & 62.5 & 65.5 & 67.6 & 66.6\\	
	%DistrEn-512(123)   & 76.3  & 60.1 & 63.8 & 60.4 & 65.1 & 67.3 & 66.2\\	
	%DistrEn-512(23)    & 76.8  & 64.8 & 63.6 & 63.2 & 65.5 & 67.7 & 66.6\\	
%DistrScaleEn(23-23)    & 77.3  & 65.6 & 63.8 & 63.7 & 65.8 & 67.9 & 66.8\\
%DistrScaleEn(23-123)   & 77.3  & 65.5 & 63.7 & 64.5 & 65.7 & 67.9 & 66.8\\
				\cline{2-11}
				&\multirow{9}{*}{R} &\multirow{3}{*}{M1} 
				  & 384 & 78.4  & 65.3 & 65.1 & 64.6 & 66.7 & 68.9 & 67.8\\ 
				&&& 448 & 79.3  & 63.0 & 66.2 & 64.1 & 66.9 & 69.1 & 68.0\\	
				&&& 512 & 79.0  & 65.9 & 65.5 & 65.1 & 67.1 & 69.3 & 68.1\\
	   %ScaleEn-M1(123) & 80.7  & 65.5 & 66.6 & 65.8 & 67.9 & 70.2 & 69.0\\ 
	   %ScaleEn-M1(23)  & 80.6  & 65.0 & 66.7 & 65.6 & 67.8 & 70.0 & 68.9\\
				\cline{3-11}	
				&& \multirow{3}{*}{M2}
				  & 384 & 79.8  & 68.9 & 66.2 & 65.7 & 67.2 & 69.4 & 68.2\\ 
				&&& 448 & 80.7  & 69.4 & 67.0 & 66.4 & 67.7 & 69.9 & 68.8\\	
				&&& 512 & 80.9  & 69.6 & 66.9 & 68.4 & 67.9 & 70.2 & 69.0\\
	   %ScaleEn-M2(123) & 81.5  & 70.5 & 67.2 & 67.9 & 68.1 & 70.4 & 69.2\\	   
	   %ScaleEn-M2( 23) & 81.4  & 70.1 & 67.2 & 67.8 & 68.1 & 70.4 & 69.2\\	   
				\cline{3-11}	
				&& \multirow{3}{*}{M3}
				  & 384 & 79.9  & 66.3 & 67.1 & 65.5 & 67.2 & 69.4 & 68.3\\ 
				&&& 448 & 81.1  & 66.9 & 67.7 & 66.0 & 67.8 & 70.0 & 68.9\\	
				&&& 512 & 81.3  & 68.1 & 67.7 & 66.5 & 67.9 & 70.1 & 69.0\\
	   %ScaleEn-M3(123) & 82.2  & 68.7 & 68.3 & 67.3 & 68.4 & 70.6 & 69.5\\ 
	   %ScaleEn-M3(23)  & 82.2  & 68.7 & 68.2 & 67.1 & 68.4 & 70.6 & 69.5\\	   
   %TrialEn-M1-384(123) & 80.3  & 65.7 & 66.7 & 65.4 & 67.7 & 69.9 & 68.7\\
   %TrialEn-M1-448(23)  & 79.8  & 65.9 & 66.3 & 65.3 & 67.4 & 69.6 & 68.5\\
   %TrialEn-M1-512(23)  & 79.8  & 66.2 & 66.3 & 65.5 & 67.5 & 69.7 & 68.5\\
   %DistrEn-448(123)    & 82.0  & 69.7 & 67.8 & 66.8 & 68.5 & 70.7 & 69.6\\
   %DistrEn-448(23)     & 82.2  & 69.8 & 68.2 & 67.5 & 68.5 & 70.7 & 69.6\\
   %DistrEn-512(123)    & 82.3  & 69.6 & 67.8 & 68.8 & 68.6 & 70.8 & 69.7\\
   %DistrEn-512(23)     & 82.4  & 70.4 & 68.0 & 69.4 & 68.6 & 70.8 & 69.7\\
%DistrScaleEn(23-23)    & 82.8  & 71.0 & 68.4 & 68.9 & 68.8 & 71.1 & 70.0\\
%DistrScaleEn(23-123)   & 82.8  & 70.6 & 68.4 & 68.1 & 68.8 & 71.1 & 69.9\\

				\hline
				\hline		
						
				\multirow{18}{*}{VOC} 
				&\multirow{9}{*}{V} &\multirow{3}{*}{M1}  
				  & 384  & 89.1 & 40.0 & 92.1 & 54.9 & 44.1 & 93.5 & 59.9\\
				&&& 448  & 89.3 & 39.3 & 92.1 & 54.2 & 44.1 & 93.4 & 59.9\\
				&&& 512  & 89.2 & 39.9 & 91.9 & 54.7 & 44.1 & 93.4 & 59.9\\
	    %ScaleEn-M1(123) & 89.7 & 40.0 & 92.4 & 54.9 & 44.3 & 93.8 & 60.1\\ 
	    %ScaleEn-M1( 23) & 89.6 & 39.8 & 92.4 & 54.7 & 44.2 & 93.7 & 60.1\\
				\cline{3-11}	
				&& \multirow{3}{*}{M2}
				  & 384 & 89.3 & 45.4 & 91.8 & 59.5 & 44.2 & 93.7 & 60.1\\ 
				&&& 448 & 89.6 & 45.3 & 92.3 & 59.5 & 44.3 & 93.9 & 60.2\\	
				&&& 512 & 89.3 & 44.8 & 92.1 & 59.1 & 44.3 & 93.8 & 60.2\\
	    %ScaleEn-M2(123)& 89.9 & 45.6 & 92.5 & 59.8 & 44.4 & 94.2 & 60.4\\	     
		%ScaleEn-M2( 23)& 89.7 & 45.2 & 92.3 & 59.4 & 44.4 & 94.1 & 60.3\\		
				\cline{3-11}	
				&& \multirow{3}{*}{M3}
				  & 384 & 89.9 & 41.0 & 92.9 & 56.2 & 44.5 & 94.2 & 60.4\\ 
				&&& 448 & 90.0 & 40.5 & 92.8 & 55.7 & 44.5 & 94.3 & 60.5\\	
				&&& 512 & 90.2 & 42.0 & 92.8 & 57.0 & 44.4 & 94.2 & 60.4\\
	    %ScaleEn-M3(123)& 90.5 & 41.4 & 93.1 & 56.6 & 44.6 & 94.5 & 60.6\\	    
		%ScaleEn-M3( 23)& 90.4 & 41.2 & 93.1 & 56.5 & 44.6 & 94.6 & 60.7\\				
   %DistrEn-384(123)    & 90.1 & 41.8 & 92.8 & 56.8 & 44.4 & 94.2 & 60.4\\   
   %DistrEn-384(23)     & 90.4 & 43.4 & 93.0 & 58.4 & 44.6 & 94.5 & 60.6\\      
   %DistrEn-448(123)    & 90.4 & 41.2 & 93.1 & 56.3 & 44.5 & 94.3 & 60.5\\   
   %DistrEn-448(23)     & 90.6 & 43.0 & 93.3 & 58.1 & 44.7 & 94.6 & 60.7\\   
   %DistrEn-512(123)    & 90.1 & 41.8 & 92.5 & 56.7 & 44.4 & 94.0 & 60.3\\   
   %DistrEn-512(23)     & 90.3 & 43.6 & 92.5 & 58.4 & 44.4 & 94.1 & 60.3\\   
%DistrScaleEn(23-23)    & 90.7 & 43.3 & 93.3 & 58.4 & 44.7 & 94.7 & 60.7\\
%DistrScaleEn(23-123)   & 90.4 & 41.6 & 93.1 & 56.7 & 44.6 & 94.4 & 60.5\\
				\cline{2-11}
				&\multirow{9}{*}{R} &\multirow{3}{*}{M1} 
				  & 384 & 93.4 & 40.5 & 94.8 & 55.9 & 45.1 & 95.6 & 61.3\\ 
				&&& 448 & 94.1 & 40.8 & 95.5 & 56.3 & 45.5 & 96.3 & 61.8\\	
				&&& 512 & 94.2 & 41.4 & 95.4 & 56.7 & 45.5 & 96.3 & 61.8\\
	    %ScaleEn-M1(123)& 94.5 & 41.2 & 95.7 & 56.7 & 45.5 & 96.5 & 61.9\\	    	    
		%ScaleEn-M1( 23)& 94.5 & 41.3 & 95.7 & 56.7 & 45.6 & 96.5 & 61.9\\		
				\cline{3-11}	
				&& \multirow{3}{*}{M2}
				  & 384 & 92.4 & 44.9 & 94.1 & 60.1 & 45.1 & 95.5 & 61.2\\ 
				&&& 448 & 92.7 & 44.9 & 94.0 & 60.0 & 45.1 & 95.5 & 61.2\\	
				&&& 512 & 92.9 & 45.6 & 94.7 & 60.7 & 45.3 & 96.0 & 61.6\\
	    %ScaleEn-M2(123)& 93.3 & 45.6 & 94.6 & 60.7 & 45.3 & 96.0 & 61.5\\	    	    
		%ScaleEn-M2( 23)& 93.2 & 45.6 & 94.7 & 60.8 & 45.3 & 96.0 & 61.6\\			
					
				\cline{3-11}	
				&& \multirow{3}{*}{M3}
				  & 384 & 93.2 & 42.0 & 94.8 & 57.5 & 45.2 & 95.8 & 61.5\\ 
				&&& 448 & 93.6 & 42.0 & 95.1 & 57.4 & 45.3 & 96.1 & 61.6\\	
				&&& 512 & 93.8 & 42.3 & 95.6 & 57.7 & 45.5 & 96.4 & 61.8\\
	    %ScaleEn-M3(123)& 94.3 & 42.7 & 95.6 & 58.1 & 45.5 & 96.4 & 61.8\\	    	    
		%ScaleEn-M3( 23)& 94.2 & 42.6 & 95.6 & 58.0 & 45.5 & 96.4 & 61.8\\
   %DistrEn-384(123)    & 94.2 & 42.8 & 95.1 & 58.3 & 45.4 & 96.1 & 61.6\\      
   %DistrEn-384(13)     & 94.1 & 41.5 & 95.3 & 57.0 & 45.4 & 96.2 & 61.7\\     
   %DistrEn-448(123)    & 94.6 & 43.0 & 95.7 & 58.6 & 45.6 & 96.6 & 62.0\\      
   %DistrEn-448(13)     & 94.6 & 41.7 & 95.8 & 57.2 & 45.6 & 96.5 & 61.9\\    
   %DistrEn-512(123)    & 94.7 & 43.3 & 95.8 & 58.9 & 45.7 & 96.8 & 62.1\\      
   %DistrEn-512(13)     & 94.7 & 42.0 & 95.8 & 57.5 & 45.6 & 96.6 & 62.0\\      
%DistrScaleEn(123-23)   & 94.8 & 43.3 & 96.0 & 59.0 & 45.7 & 96.9 & 62.1\\
%DistrScaleEn(123-123)  & 94.8 & 43.3 & 95.9 & 59.0 & 45.7 & 96.7 & 62.1\\
						
		\hline
				
			\end{tabular}
		\end{lrbox}
		\scalebox{0.62}{\usebox{\tablebox}}
	}
\end{table}

%------------------------------------------------------------------------- 
 As described in Section \ref{sect:ensemble}, we employ \textit{multi-scale ensemble (ScaleEn)} and \textit{distribution ensemble (DistrEn)}. For multi-scale ensemble, three M3 scores are fused since it generally performs better than M1 and M2 except on VOC dataset when Resnet101 is used for which M1 scores are fused for the best performance. For distribution ensemble, we choose M2 and M3 learned with the image size of 512. Results are shown in Table \ref{table:resCmp}. It is obvious that both types of ensembles can achieve better performance than our best single model. Note that the results of \textit{ScaleEn} and \textit{DistrEn} are based on the score fusion of three and two models respectively, and a fusion of more models would lead to better results.

We also compare the proposed baseline performance against that of state-of-the-art approaches including RCP \cite{wang2016beyond} which uses a random cropping pooling layer capturing multi-scale information, WILDCAT \cite{durand2017wildcat} which designs novel class-wise and spatial pooling strategies as well as employs multi-scale input images, RLSD \cite{zhang2018multi} which exploits the label dependencies using a CNN-RNN framework, AttRegion \cite{wang2017multi} and ResNet-SRN-att \cite{zhu2017learning} employing attention mechanisms in their models. Without these tricks, we only use the basic deep models and score-level fusion but achieve better performance on NUS-WIDE (e.g., 59.3\% vs 54.1\% mAP and 62.0\% vs 60.5\% overall $F_1$ when using VGG16) and MS-COCO (e.g., 76.8\% vs 67.4\% mAP when using VGG16 and 82.4\% vs 80.7\% mAP when using ResNet101) datasets, comparable performance on VOC2007 when ResNet101 is used (e.g., 94.7\% vs 95.0\% mAP) as indicated by the \textbf{bold} font in Table \ref{table:resCmp}. As a result, our experimental results indicate that the basic deep models with proper training strategies have more capabilities than what has been explored for multi-label image classification and a strong baseline is presented.

\begin{table}[!htbp]
	\centering
	{%\centering
		\caption[]{Comparison with state-of-the-art results on three benchmark datasets. \small (Notations are the same as those in Table \ref{table:main}. For a fair comparison, we do not list threshold based precision/recall/$F_1$ reported in literature \cite{wang2017multi,ge2018multi}.)
		}
		\label{table:resCmp}
		\begin{lrbox}{\tablebox}
			\begin{tabular}{c|c|c|c|ccc|ccc}
				\hline
				\textbf{DS}&\textbf{BM}&\textbf{Method} & \textbf{mAP}  & \textbf{L-P} & \textbf{L-R} & \textbf{L-F$_1$} & \textbf{O-P} & \textbf{O-R} & \textbf{O-F$_1$}\\ 
				
				\hline
				\multirow{11}{*}{NUS} 
				&V & CNN-RNN \cite{wang2016cnn} & -&  40.5 & 30.4  & 34.7 & 49.9 & 61.7 & 55.2 \\
				&V& RLSD \cite{zhang2018multi} & 54.1 & 44.4 & 49.6 & 46.9 & 54.4 & 67.6 & 60.3 \\
				&V&WARP \cite{li2017improving}&- & 43.8& {57.1}&- & 54.5&67.9&60.5\\	
				%\cline{3-10}
				&V& Single Best& 59.0 &45.7 & 55.1 & 46.6 & 55.9 & 68.9 & 61.7\\	
				&V& ScaleEn & 59.1 &47.2 & 54.9 & 47.2 & 56.1 & 69.0 & 61.9\\
				&V& DistrEn & \textbf{59.3} &47.0 & 55.0 & \textbf{47.0} & 56.2 & 69.1 & \textbf{62.0}\\
				\cline{2-10}
				&{R} &ResNet-SRN-att \cite{zhu2017learning} & 61.8 & 47.4 & 57.7 & 47.7 & 56.2 & 69.6 & 62.2\\
				&{R} &ResNet-SRN \cite{zhu2017learning} & \textbf{62.0} & 48.2 & 58.9 & 48.9 & 56.2 & 69.6 & 62.2\\
				%\cline{3-10}
				&R& Single Best & 60.8 &45.8 & 60.6 & 49.2 & 56.2 & 69.2 & 62.0\\ 
				&R& ScaleEn& 61.7 &46.9 & 60.5 & 49.9 & 56.7 & 69.7 & 62.5\\
				&R& DistrEn & \textbf{62.0} &46.8 & 61.1 & \textbf{49.9} & 56.7 & 69.8 & \textbf{62.6}\\
				%&V+R& Ours-En3 \\
				\hline
				\hline
				\multirow{11}{*}{COCO}
				&V& WARP \cite{li2017improving} & - & 55.5 & 57.4 &- & 59.6 & 61.5& 60.5\\
				&V& Ranking \cite{li2017improving} & - & 57.0 & 57.8 & - & 60.2 & 62.2 & 61.2\\
				&V& RLSD \cite{zhang2018multi} & 67.4 & - & - & -& -& - &- \\
				%\cline{3-10}
				&V& Single Best & 75.9  & 64.3 & 62.6 & 63.4 & 65.0 & 67.1 & 66.1\\	 
				&V& ScaleEn& 76.5  & 65.2 & 63.0 & \textbf{64.0} & 65.4 & 67.5 & 66.4\\
				&V& DistrEn & \textbf{76.8}  & 64.8 & 63.6 & 63.2 & 65.5 & 67.7 & \textbf{66.6}\\	
				\cline{2-10}
				&R&ResNet-SRN \cite{zhu2017learning} & 77.1  & - & - &-& - & - & -  \\
				&R&WIDECAT \cite{durand2017wildcat} &80.7 &- & - &- & - & -&- \\
				&R& Single Best& 81.3  & 68.1 & 67.7 & 66.5 & 67.9 & 70.1 & 69.0\\
				&R& ScaleEn & 82.2  & 68.7 & 68.3 & 67.3 & 68.4 & 70.6 & 69.5\\ 
				&R& DistrEn & \textbf{82.4}  & 70.4 & 68.0 & \textbf{69.4} & 68.6 & 70.8 & \textbf{69.7}\\
				\cline{3-10}
				
				\hline
				\hline
				\multirow{11}{*}{VOC}&V &  CNN-RNN \cite{wang2016cnn} & 84.0&  - & -  & - & - & - & - \\
				&V& AttRegion\cite{wang2017multi} & 91.9 & - & - & - & - & - & -\\
				&V& RLSD \cite{zhang2018multi} & 87.3 & {50.5} & {90.6} & \textbf{64.9} & {47.5}& {92.4} &\textbf{62.7} \\
				&V& RCP \cite{wang2016beyond} & \textbf{92.5} &- & -&-&-&-&-\\
				%\cline{3-10}
				&V& Single Best& 90.2 & 42.0 & 92.8 & 57.0 & 44.4 & 94.2 & 60.4\\
				&V& ScaleEn & 90.5 & 41.4 & 93.1 & 56.6 & 44.6 & 94.5 & 60.6\\
				&V& DistrEn & 90.6 & 43.0 & 93.3 & 58.1 & 44.7 & {94.6} & 60.7\\   
				\cline{2-10}
				&R& WILDCAT \cite{durand2017wildcat}& \textbf{95.0} &-&-&-&-&-&-\\
				&R& Single Best & 94.2 & 41.4 & 95.4 & 56.7 & 45.5 & 96.3 & 61.8\\ 
				&R& ScaleEn& 94.5 & 41.2 & 95.7 & 56.7 & 45.5 & 96.5 & 61.9\\	    	    
				&R& DistrEn & 94.7 & 42.0 & 95.8 & \textbf{57.5} & 45.6 & 96.6 & \textbf{62.0}\\     
				\hline
				
			\end{tabular}
		\end{lrbox}
		\scalebox{0.62}{\usebox{\tablebox}}
	}
\end{table}

%===========================================================
\section{Conclusion}

In summary, we investigate the impacts of varying input image sizes and data augmentation techniques in multi-label image classification and present a simple yet effective score level fusion to explore the complementarity of different learned models, achieving state-of-the-art performance on three benchmark datasets. The results of extensive experiments presented in this paper demonstrate a proper exploration of multi-scale information and data augmentation techniques will benefit multi-label image classification hence should be considered when designing new deep architectures for multi-label image classification in future studies. 
% Emphasize the contributions again....

\bibliographystyle{IEEEbib}
\small
\bibliography{refs}

\end{document}